\documentclass[runningheads]{llncs}

\usepackage{eccv}

\usepackage{eccvabbrv}

\usepackage{graphicx}
\usepackage{booktabs}

\usepackage[accsupp]{axessibility}  

\usepackage{hyperref}

\usepackage{orcidlink}

\begin{document}

\title{MM-Conv: A Multi-Modal Conversational Dataset for Virtual Humans} 


\author{Anna Deichler \orcidlink{0000-0003-3135-5683} \and
Jim O'Regan \orcidlink{0000-0003-2598-6868} \and
Jonas Beskow\orcidlink{0000-0003-1399-6604 }}

\authorrunning{A.~Deichler et al.}

\institute{KTH Royal Institute of Technology, Stockholm, Sweden}

\maketitle

\begin{abstract}
In this paper, we present a novel dataset captured using a VR headset to record conversations between participants within a physics simulator (AI2-THOR). Our primary objective is to extend the field of co-speech gesture generation by incorporating rich contextual information within referential settings. Participants engaged in various conversational scenarios, all based on referential communication tasks. The dataset provides a rich set of multimodal recordings such as motion capture, speech, gaze, and scene graphs. This comprehensive dataset aims to enhance the understanding and development of gesture generation models in 3D scenes by providing diverse and contextually rich data.
\end{abstract}

\section{Introduction}
\label{sec:intro}
Referential Communication is a specific mode of communication that often occurs within a situated dialogue. This can include identifying, describing, or giving instructions related to objects, locations, or people. This form of communication bridges the perceptual and conceptual understanding of one's surroundings. This relies on multimodal expressions, including spatial language and non-verbal behaviors like gaze and pointing gestures. When discussing spatial contexts, pointing or gesturing becomes a crucial addition to spatial language, providing a more immediate and often clearer method of specifying locations or directing attention to particular objects or areas. For agents to effectively understand and participate in referential communication within a situated dialogue, they need to be capable of interpreting and generating both verbal spatial references and non-verbal cues such as pointing gestures and gaze. This dual capability allows for a more nuanced and efficient exchange of information. 

Non-verbal behavior generation in virtual humans has evolved significantly. Recently, data-driven solutions have become prominent, with supervised learning systems achieving high naturalness in certain settings. This advancement was enabled by the release of large-scale human motion datasets. One specific sub-field of human motion generation is co-speech gesture generation. The availability of large, synchronized speech (audio and text) and motion gesture datasets \cite{ferstl2021expressgesture},\cite{lee2019talking} has allowed significant enhancements in motion naturalness but it has also narrowed down the behavior scope to isolated co-speech gesture generation.

Our proposed dataset aims to introduce a novel perspective on gesture generation research, focusing on data-driven situated gesture generation by introducing a new dataset. Currently, there is a lack of datasets that enable the generation of gestures in situated settings. We propose the integration of virtual reality (VR) in gesture generation research, as it offers a transformative approach to studying the intricacies of spatial language with greater experimental control.  We have recorded 6.7 hours of synchronized speech, motion, facial expression, and gaze data, alongside scene graphs in a VR and motion capture-based system. The contributions of the dataset could also be understood in a broader setting, as it includes a rich set of human verbal and non-verbal communication modes, such as speech, gaze, facial expressions, and pointing gestures. It allows for the study of how individuals communicate spatial information in task-oriented settings and can be leveraged to reproduce these behaviors in agents.
\section{Related Work}
\label{sec:related_work}
\subsection{Co-Speech Gesture Generation}
The generation of co-speech gestures has seen significant advances through data-driven methods that create natural motion patterns. However, these approaches often generate gestures in isolation, lacking spatial context. Most of these recent systems are based on generative models that take audio (and sometimes text) as input conditioning information \cite{ao2023gesturediffuclip} \cite{deichler2023diffusion}, and are limited to generating beat gestures aligned with speech.
\subsection{Human-Scene Interaction Datasets}
Human-scene interaction datasets are crucial for creating realistic virtual agents.  Most prior work on aligning human motion sequences with 3D scenes has focused on locomotion and human-object interaction tasks \cite{wang2022humanise}. 
VR and AR-based datasets, such as Nymeria \cite{ma2024nymeria}, provide rich multimodal data, capturing human motion and interactions within various scenarios. These datasets have been instrumental in advancing research in areas like egocentric motion capture and action recognition. The inclusion of VR devices for data collection enables more immersive and realistic interactions, providing a valuable resource for developing and testing AI models.

Integrating environmental cues into gesture generation models by leveraging these two distinct fields would allow for more natural and effective human-agent communication.
\section{Dataset Description}
We combine motion capture with virtual reality to record conversations in spatial settings using VR headset and the AI2-THOR \cite{ai2thor} physics simulator. The motivation for this setup is twofold: first, VR provides us with experimental control over the environment, allowing us to manipulate and replicate conditions precisely. Second, it simplifies the annotation of objects and scenes, making the data more structured and easier to analyze.
The dataset captures multiple, time-synchronized modalities to provide a comprehensive view of the interactions:
\begin{itemize}
    \item  Motion capture: Full-body (50-marker skeleton) motion is recorded to capture detailed gesture and movement data for both the main speaker and interlocutor.
    \item Speech: Audio recordings of the conversations for both speakers.
    \item Gaze: Eye-tracking data is recorded with the headset to understand visual attention and focus of the main speaker, synchronized to Motive's timecode.
    \item Facial expressions: 52 blend shapes are recorded with the headset, synchronized to Motive's timecode.
    \item Scene graphs: Object placements are annotated and provided in JSON format for each room.
\end{itemize}
For additional details, please refer to the official dataset webpage
\footnote{\url{https://mm-conv.github.io/}}.

\begin{table}[tb]
  \caption{Total recording time and average take length (in minutes) for each speaker, with the reported values representing the mean ± standard deviation.
  Speaker IDs match the convention in the dataset.
  }
  \label{tab:times}
  \centering
  \begin{tabular}{@{}lll@{}}
    \toprule
    Speaker ID & Total time (h)  & Avg. take length (mn) \\
    \midrule
    3  & $1.045$ & $7.83 \pm 1.56$\\
    4 & $1.116$ & $5.15 \pm 1.52$ \\
    5 & $1.746$ & $8.73 \pm 2.32$\\
    6 & $1.4$ & $7.62 \pm 1.3$\\
    7 & $1.39$ & $7.56 \pm 2.22$\\
        \hline
    total & 6.7 &  $7.30 \pm 2.19$ \\
    
  \bottomrule
  \end{tabular}
\end{table}

\subsection{Data Collection}
\begin{figure}[tb]
  \centering
  \begin{subfigure}{0.3\textwidth}  
    \centering
    \includegraphics[width=\textwidth]{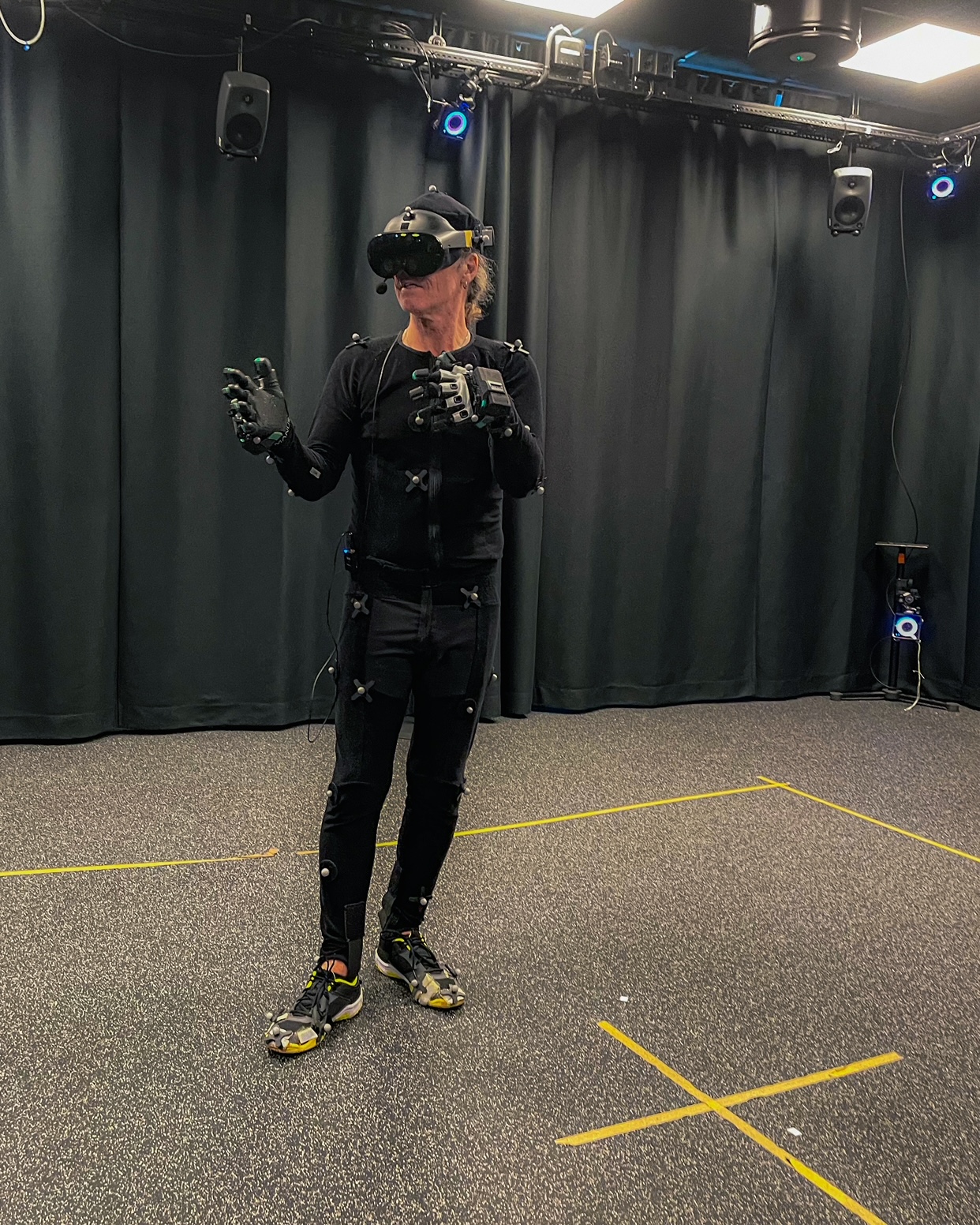}  
    \caption{Illustration of hardware setup.}
    \label{fig:setup1}
  \end{subfigure}
  \hspace{0.03\textwidth}  
  \begin{subfigure}{0.6\textwidth}  
    \centering
    \includegraphics[width=\textwidth]{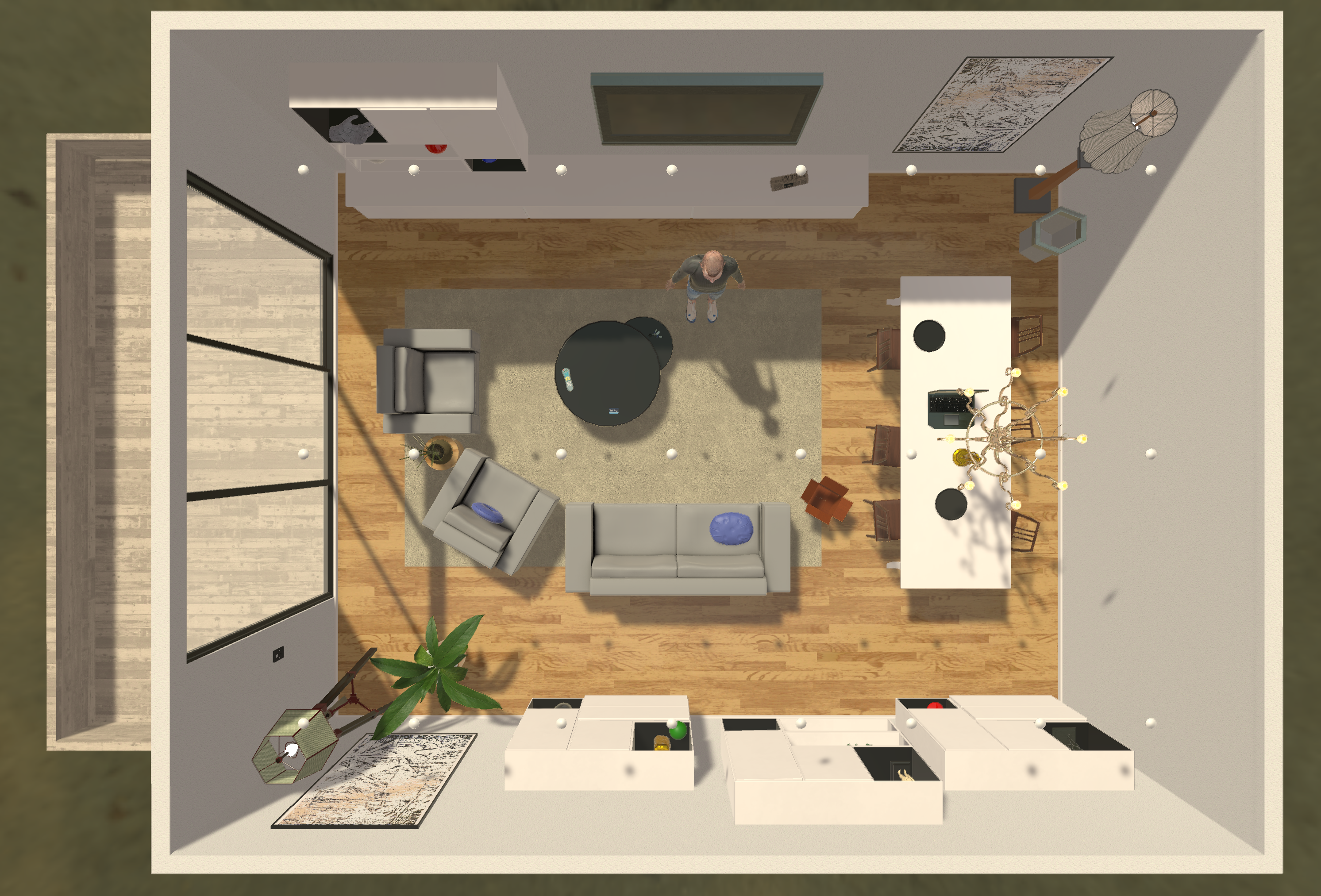}  
    \caption{Example scene from simulator.}
    \label{fig:setup2}
  \end{subfigure}
  \caption{(a) Illustrations of the hardware setup as described in Appendix \ref{app:hardware}. The participants wear a motion capture suit with gloves motion tracking and a VR headset with gaze tracking in the experimental setup. (b) Top-down view of one of the scenes used in the experiments.}
  \label{fig:setup}
\end{figure}

We invite 5 participants (main speaker) to interact with an experimenter (interlocutor) in spatial description scenarios, tailored for the data collections' aims. Both participants' motions were tracked, but only the main actor wore a gaze-tracking device. The cleaned total recording time is 6.7 hours, further per-speaker information can be found in Table \ref{tab:times}. Prior to recording, we chose 5 rooms from the AI2-THOR simulator to run the experiments. The average number of "interactable" objects in these rooms is $38 \pm 3.16$. Further information about object distribution can be found in Appendix \ref{app:objects}. We record 2-3 scenarios in each room for each participant. The emphasis on the main actor (wearing the headset) who introduces or discusses objects in a room in different scenarios. The second actor reacts to the main actor (does not introduce new objects). Further information about the scenarios can be found in the Appendix \ref{app:scen}. Our data capture system consists of external motion capture tracking, finger-tracking gloves, and a gaze-tracking VR headset. Figure \ref{fig:setup} shows an example of this setup and an example scene from the simulator. Further details about the hardware setup can be found in Appendix \ref{app:hardware}.



\subsection{Data Annotation}
In the data annotation process, all modalities were synchronized using the shared timecode information. Due to the nature of the data, the audio transcription required an elaborate process, which is detailed in Appendix  \ref{app:text}. Finally, the scene graphs were extracted from the simulator in JSON format.

\section{Conclusions}
The dataset presented in this paper offers a comprehensive resource for training and testing gesture generation models in situated settings. Recent advances in diffusion-based human motion generation models, such as augmented control approaches like OmniControl \cite{xie2024omnicontrol} and TESMO\cite{yi2024tesmo}, have introduced new methods for enhancing scene-aware motion generation by incorporating spatial control signals. 
When combined with these methods, the referential dataset could help enable the development of more sophisticated gesture generation models that are responsive to their environment. By providing rich, synchronized multimodal data, including motion capture, speech, gaze, and scene graphs, this dataset enhances our understanding of human communication in task-oriented scenarios and can significantly contribute to the development of more natural and contextually aware virtual agents.

\bibliographystyle{splncs04}
\bibliography{main}
\section{Appendix}
\appendix

\section{Scenarios}
In our study, five different rooms were selected from the AI2-THOR simulator. Across these rooms, participants engaged in 2-3 scenarios per room. This replication across rooms allowed us to maintain consistent interaction patterns while observing the influence of different spatial environments on participant behavior and ensuring that the dataset contains a distribution of objects at different distances and in different directions. The scenarios were the following:
\label{app:scen}

\paragraph{General Setup:}
\begin{itemize}
    \item Location: A room in an apartment.
    \item Duration: Each scene lasts approximately 5-7.5 minutes.
    \item Focus: Emphasis on the main actor who introduces or discusses objects in a room in different scenarios. The second actor reacts to the main actor (does not introduce new objects).
\end{itemize}

\paragraph{Scenario 1}: Bragging/Introducing New Apartment

Main Actor: Apartment Owner,
Secondary Actor: Friend

Description: The apartment owner shows off their new apartment to their friend, highlighting various features and objects in the room.

Dialogue Focus: Description of the objects, personal anecdotes about their acquisition, and the benefits of each item.

\paragraph{Scenario 2}: Landlord Asking About Objects

Main Actor: Landlord, 
Secondary Actor: Tenant

Description: The landlord inquires about various objects in the apartment, possibly checking for maintenance needs or understanding the tenant’s living conditions.

Dialogue Focus: Questions about the objects, their usage, condition, and any issues.

\paragraph{Scenario 3}: Interior Designer Giving Tips

Main Actor: Interior Designer, 
Secondary Actor: Client

Description: The interior designer provides suggestions and advice on improving the apartment’s aesthetics and functionality.

Dialogue Focus: Suggestions for rearranging furniture, adding new decor items, and making the space more efficient.

\clearpage
\section{Object distribution}
\begin{figure}[htb]
  \centering
  \includegraphics[height=9cm]{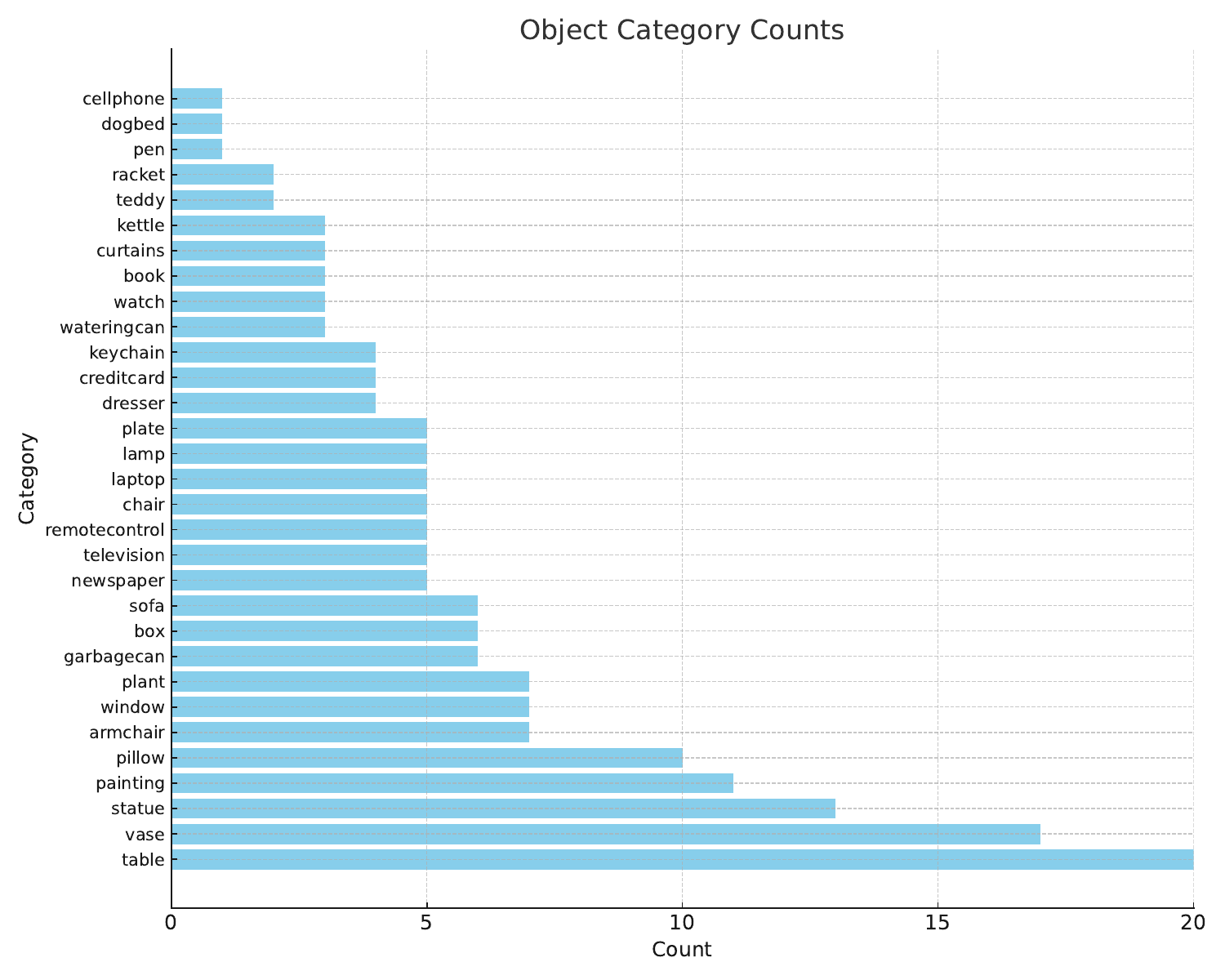}
  \caption{Object category distribution across simulated environments used in the experiments.
  }
  \label{fig:object}
\end{figure}

Figure \ref{fig:object} shows the distribution of object categories in all rooms.
\label{app:objects}

\section{Hardware setup details}
\label{app:hardware}
Skeletal information is recorded using the Optitrack Motive \cite{optitrack} motion capture system.  An OptiTrack motion capture system comprising 16 Prime x 41 cameras. This system tracks the movements of participants using 50 marker-based skeletons, which are redefined for every actor to ensure accurate motion capture. 

Accurate finger movements are captured using Quantum Mocap Metagloves \cite{manus}. The full-body skeletal information from the Optitrack system and the finger-tracking information from the Manus Gloves are synchronized within the Optitrack system to ensure comprehensive motion data. Tentacle Sync E \cite{tentacle} devices are used to generate time codes, which is used to synchronize audio track and the Optitrack mocap system. All devices are managed centrally during recording. A coordinated start/stop functionality provides coarse synchronization within hundreds of milliseconds. We use SMPTE time code for fine-grained synchronization.

This time code is streamed to the AI2-THOR simulator and recorded alongside the face and gaze tracking data from the META Quest Pro \cite{quest} headset. Additionally, the position information of the Meta Quest headset is adjusted to the motion capture head position before rendering the simulation, in order to achieve synchrony between the two distinct systems.

\section{Text annotation}
\label{app:text}
To obtain segment level transcriptions of the audio, we used whisperX~\cite{bain2022whisperx}, which employs a voice activity detection module to obtain better timestamps. For better accuracy, we prompted whisperX with in-domain text that had been edited to include filled pauses, repeated words, and number words instead of digits. Although this improved the accuracy of the transcription, we still needed to post-process the results to replace remaining digits. Although whisperX can produce word-level timestamps, they were insufficiently accurate for our purposes. To obtain these timestamps, we used the tools created for the librispeech corpus~\cite{panayotov2015librispeech} to intersect whisperX’s output with the output from a system based on Kaldi~\cite{Povey_ASRU2011}, taking the timestamps from that system, while accepting any insertions of a set of filled pauses regardless of context, and other words if the word to either side was the same. In the event of remaining disagreements between the systems, we used the Montreal Forced Aligner~\cite{mcauliffe17_interspeech} to force alignment to whisperX’s output.

\end{document}